\documentclass{article}

\usepackage{neurips_2024}

\usepackage[utf8]{inputenc} 
\usepackage[T1]{fontenc}    
\usepackage{url}            
\usepackage{booktabs}       
\usepackage{amsfonts}       
\usepackage{nicefrac}       
\usepackage{microtype}      
\usepackage{xcolor}         

\usepackage{amsmath,amssymb}
\usepackage{algorithmic}
\usepackage{graphicx}
\usepackage[ruled,vlined]{algorithm2e}
\usepackage{comment}
\usepackage{caption}
\usepackage{subcaption}
\usepackage{tabularx}
\usepackage{multirow}
\usepackage{makecell}
\usepackage{placeins} 

\usepackage[backend=biber,style=numeric,sorting=none]{biblatex}
\title{A Distributed Training Architecture For Combinatorial Optimization}

\author{%
  Yuyao Long \\
  Independent Researcher \\
  \texttt{yyl15872183526@gmail.com} \\
}

\begin{document}

\maketitle

\begin{abstract}
In recent years, graph neural networks (GNNs) have been widely applied in tackling combinatorial optimization problems. However, existing methods still suffer from limited accuracy when addressing that on complex graphs and exhibit poor scalability, since full training requires loading the whole adjacent matrix and all embeddings at a time, the it may results in out of memory of a single machine. This limitation significantly restricts their applicability to large-scale scenarios. To address these challenges, we propose a distributed GNN-based training framework for combinatorial optimization. 
In details, firstly, large graph is partition into several small subgraphs. Then the individual subgraphs are full trained, providing a foundation for efficient local optimization. Finally, 
reinforcement learning (RL) are employed to take actions according to GNN output, to make sure the restrictions between cross nodes can be learned. Extensive experiments are conducted on both real large-scale social network datasets (e.g., Facebook, Youtube) and synthetically generated high-complexity graphs, which demonstrate that our framework outperforms state-of-the-art approaches in both solution quality and computational efficiency. 
Moreover, the experiments on large graph instances also validate the scalability of the model.
\end{abstract}

\section{Introduction}
Combinatorial optimization is a fundamental interdisciplinary area of operations research and computer science, focusing on finding optimal solutions within discrete search spaces, which can be applied in complex real-world decision-making. 
For example, optimizing vehicle routing can reduce costs of  and enhances efficiency in delivering passengers  \cite{Johnson2022}; in chip design,  wiring optimization can minimize interference to improve performance \cite{Lee2021}; in Internet services, refined recommendation ranking 
can inprove the experience of user \cite{Wang2020}; and in energy systems, it also can be applied to optimize power grid load scheduling balances supply and demand \cite{Brown2023}. 
As data scales and problem complexity continue to grow, the combinatorial optimization methods face increasing limitations in efficiency and adaptability because that most of the combinatorial problems are NP hard, and as the size of problem grows, the cost time will increase exponentially. On the other hand, the growing size is also a challenge for a single computing node, as it should be stored in memory when do the computing.
Consequently, obtaining high-quality approximate solutions under constrained computational resources has become essential to overcoming these bottlenecks, motivating research into more efficient and intelligent solution paradigms \cite{Liu2024}. 

Traditional algorithms for combinatorial optimization have achieved significant success through decades of development. 
The existing methods such as integer programming and dynamic programming, as well as heuristic approaches like genetic algorithms and simulated annealing \cite{Tahami2022,Raidl2008}, 
have displayed strong performance in solving combinatorial optimization problems, and widely applied in optimal decision-making, such as logistics scheduling and production planning. However, unlike traditional algorithmic design, which requires manual encoding of problem constraints and often loses intrinsic data dependencies, training neural networks to emulate the reasoning processes of classical algorithms can develope an end-to-end pipeline from raw data to decisions, which can reduce the information loss caused by manual preprocessing \cite{Cappart2021}. In recent years, Graph Neural Networks (GNNs) and Reinforcement Learning (RL) have shown strong potential for addressing combinatorial optimization problems. 
Especially, GNNs is designed to process graph-structured data, which can capture topological relationships and feature dependencies between nodes and edges through aggregating neighbors features. In the result of that, it can learn the inherent combinatorial constraints of the problem. This property has led to their widespread adoption in solving complex graph-based optimization tasks. Meanwhile, many classical algorithms, such as branch-and-bound—iteratively construct optimal solutions by sequentially selecting elements, a process that closely parallels reinforcement learning, where agents achieve goals by sequential decision-making. Hence, RL naturally  is another powerful paradigm for training neural networks to tackle combinatorial optimization problems. 

Despite these advances, GNN and RL approaches still face two major challenges. 
First, models often performe low solution accuracy when dealing with highly complex graph structures \cite{Gamarnik_2023}. 
Second, methods suffer from limited scalability when the nodes number of graph increases to thousands and millions \cite{Peng_Choi_Xu_2020}. 
The scalability issue in RL is further exacerbated, since its sample inefficiency and the exponential growth of the state space, making it difficult for the policy of reinforcement learning converges to the optimal one \cite{Darvariu_Hailes_Musolesi_2024}. 
At the same time, full-graph training results a major bottleneck for GNNs. Storing node embeddings, adjacency matrices, and intermediate activations for massive graphs demands enormous memory resources, which often exceeds the capacity of a single GPU \cite{Meirom_Maron_Mannor_Chechik_2020}. 
This hardware limitation constrains single-node scalability. Consequently, relying on single-node training is not enough for achieving scalability in large-scale combinatorial optimization. Distributed implementations thus naturally becomes a way to  solve memory and computational bottlenecks. 
Although Heydaribeni et al. \cite{Heydaribeni_Zhan_Zhang_Eliassi-Rad_Koushanfar_2023} had proposed a hypergraph-based distributed training approach, 
it still has some critical drawbacks. On the one hand, it still struggles with handling highly complex graphs, on the other hand, it remains limited by device memory, because it requires synchronous training across all subgraphs, which not only restricts scalability for large graphs but also introduces significant communication cost on embedding synchronization at every epoch. Moreover, it cannot adapt to dynamic graph scenarios, when there are newly added subgraphs, it should be full retraining to maintain consistency.

To address these challenges, we propose a distributed training architecture that integrates Reinforcement Learning (RL) as a coordination mechanism to fine-tune GNNs. The conceptual foundation of this framework arises from two key observations. First, the classical dynamic programming approach to the combinatorial optimization problem, for example, the Maximum Independent Size(MIS) follows a recursive decision paradigm: for each node, it compares the expected optimal MIS size between either including or excluding the node and derives the global optimal solution through state transitions. This binary decision made by iterative evaluation and taking reward as feedback is close to RL’s decision mechanism, where the Q-value function select action which can achieve the best reward state-value, and the state-value is calculated through iterative interaction with the environment. Second, for a pre-trained GNN, it has already captured topological patterns and local dependencies, and the binary inclusion or exclusion of a node does not alter the overall graph structure. 
This allows the GNN can adapt to new decision scenarios through targeted fine-tuning instead of full retraining. 

Our proposed framework consists of two core components, and these collaborate together to find optimal solutions for large-scale combinatorial optimization problems:
\begin{enumerate}
    \item \textbf{Subgraph Partitioning and Independent Training:} We adopt the Louvain algorithm \cite{Blondel_Guillaume_Lambiotte_Lefebvre_2008} for subgraph partitioning, effectively preserving intra-subgraph connectivity while minimizing cross-subgraph edges. This ensures each subgraph maintains structural integrity, reducing inter-subgraph conflicts. Each subgraph is then trained independently using a GNN, enabling parallel local optimization and significantly lowering the computational burden of full-graph training—crucial for scalability to large graphs.
    \item \textbf{Inter-Subgraph Conflict Coordination:} For nodes that connect different subgraphs and cause conflicts between local solutions, we design a Q-learning–based RL mechanism to guide optimal node selection. The RL agent triggers targeted fine-tuning of the conflicting subgraphs, with global solution improvement (e.g., MIS size or conflict reduction) serving as the reward signal. This allows the agent to learn a globally optimal coordination policy that resolves inter-subgraph conflicts while maintaining strong local solution quality.
\end{enumerate}

The remainder of this paper is organized as follows. Section II reviews prior research on RL and GNN methods for the MIS problem, including distributed training techniques for large-scale graph optimization. Section III presents the proposed distributed GNN–RL training framework, detailing the graph partitioning process, subgraph-level training, and the Q-learning–based coordination mechanism. Section IV discusses the experimental results and their implications.

\section{Related Work}

Since the introduction of the Pointer Network (Ptr-Net) by Vinyals et al.~\cite{Vinyals_Fortunato_Jaitly_2015} for solving classical combinatorial optimization problems, researchers have been inspired to explore neural network approaches for approximating traditional optimization algorithms. The Ptr-Net is trained via supervised learning using sample generation, predicting the visiting order of city indices to approximate the shortest tour. However, as the number of cities increases, obtaining exact solutions through traditional solvers becomes computationally prohibitive. To mitigate this, the authors trained the model using approximate solutions obtained from heuristic solvers, which inevitably limited the overall accuracy of learned policies.

To address the dependency on labeled data, Toenshoff et al.~\cite{Toenshoff_Ritzert_Wolf_Grohe_2019} proposed an unsupervised learning model, RUN-CSP, that reformulates constraint satisfaction problems (CSPs) as differentiable loss functions. By maximizing the probability of constraint satisfaction, RUN-CSP enables learning without explicit supervision and demonstrates general applicability to various binary CSPs. Building upon similar principles, Schuetz et al.~\cite{Schuetz_Brubaker_Katzgraber_2022} introduced the Physics-Inspired Graph Neural Network (PI-GNN), which formulates NP-hard problems such as the Maximum Independent Set (MIS) as a Quadratic Unconstrained Binary Optimization (QUBO) problem. By minimizing the QUBO energy function, PI-GNN achieves unsupervised end-to-end optimization. Compared to RUN-CSP, the message-passing mechanism in GNNs allows PI-GNN to scale more effectively to large graphs, where neighboring nodes exchange information iteratively. Experiments by Schuetz et al. showed that PI-GNN can handle graphs with hundreds of thousands of nodes. Nevertheless, both RUN-CSP and PI-GNN exhibit severe performance degradation on graphs with high average node degrees or dense constraints, where convergence and solution accuracy drop sharply.

To overcome this limitation, Brusca et al.~\cite{NEURIPS2023_7fe3170d} proposed a dynamic recursion-based framework that decomposes the MIS problem into a sequence of subgraph selection steps. By iteratively generating and comparing subgraphs—representing cases where a node is either included or excluded—the model estimates the optimal configuration recursively. While this approach significantly improves solution quality on dense graphs, it introduces substantial computational and memory overhead, making it unsuitable for large-scale instances. More recently, Tao et al.~\cite{Tao_Aihara_Chen_2024} proposed a brain-inspired chaotic loss function to enhance PI-GNN’s ability to escape local minima. Although this method simplifies training and improves convergence stability, its effectiveness has only been verified on graphs with an average degree of three, leaving performance on highly complex graphs an open question.

Beyond GNN-based methods, iterative optimization remains a fundamental idea in classical algorithms, naturally motivating the application of reinforcement learning (RL) to combinatorial optimization. Bello et al.~\cite{Bello_Pham_Le_Norouzi_Bengio_2016} integrated RL with Ptr-Net to address the limitations of approximate supervised training, optimizing network outputs via policy gradients. However, Ptr-Net’s sequential decoding process was found to be sensitive to node ordering and difficult to parallelize. Kool et al.~\cite{Kool_Hoof_Welling_2018} subsequently replaced the encoder–decoder structure with a Transformer-based attention mechanism, achieving faster training and improved generalization. Nevertheless, RL-based approaches generally require problem-specific reward modeling, which hinders general applicability, and most experiments remain limited to small-scale graphs.

For deep learning models exceeding the memory capacity of a single device, distributed computation is a common strategy. However, in combinatorial optimization, node dependencies and structural constraints make naive graph partitioning prone to information loss, often resulting in invalid or suboptimal solutions. Distributed GNN training is typically categorized into asynchronous and synchronous paradigms. In the asynchronous case, as demonstrated by Gandhi et al.~\cite{273707}, each machine independently computes partial results for its local features before aggregating them globally, while weight synchronization still requires synchronous updates despite attempts to reduce communication frequency~\cite{Niu_Recht_Re_Wright_2011,Harlap_Narayanan_Phanishayee_Seshadri_Devanur_Ganger_Gibbons_2018}. In contrast, synchronous methods involve explicit communication of matrix partitions between devices before local computation. The HyperGNN framework proposed by Heydaribeni et al.~\cite{Heydaribeni_Zhan_Zhang_Eliassi-Rad_Koushanfar_2023} employs such a synchronous strategy for large-scale combinatorial optimization, where each GPU trains a local model, computes gradients, and synchronizes updates across devices. Although this approach alleviates memory constraints, synchronization and gradient aggregation incur substantial communication overhead, making it difficult to scale further in GPU-limited or bandwidth-constrained environments.

\section{Distributed GNN Architecture}

\subsection{Louvain algorithm for graph partition}
The Louvain algorithm, a scalable community detection method, is employed for graph partitioning in this study. This heuristic approach efficiently identifies tightly connected communities by optimizing modularity, with linear computational complexity relative to network size \cite{Blondel_Guillaume_Lambiotte_Lefebvre_2008}.

The algorithm partitions graphs by maximizing modularity gain ($\Delta Q$), defined as:
\begin{equation}
\resizebox{\columnwidth}{!}{$
\Delta Q = \left[ \frac{\Sigma_{\mathrm{in}} + k_{i,\mathrm{in}}}{2m} - \left( \frac{\Sigma_{\mathrm{tot}} + k_i}{2m} \right)^2 \right] - \left[ \frac{\Sigma_{\mathrm{in}}}{2m} - \left( \frac{\Sigma_{\mathrm{tot}}}{2m} \right)^2 - \left( \frac{k_i}{2m} \right)^2 \right]
$}
\end{equation}
where:

- $\Sigma_{\rm in}$: Sum of internal edge weights in the target community,

- $\Sigma_{\rm tot}$: Sum of edge weights connecting the target community to the rest of the graph,

- $k_{i,{\rm in}}$: Edge weight sum between node $i$ and the target community,

- $k_i$: Total edge weight degree of node $i$,

- $m$: Total edge weight sum of the entire network.

A simplified form of $\Delta Q$ clarifies the optimization logic:
\begin{equation}
\Delta Q = \frac{k_{i,\mathrm{in}}}{m} - \frac{\Sigma_{\mathrm{tot}}\, k_i}{2m^2}
\end{equation}

Modularity gain increases under two key conditions:

1. Larger $k_{i,{\rm in}}$ (strong connections between node $i$ and the target community),

2. Smaller $\Sigma_{\rm tot} \cdot k_i$ (weaker connections between the target community and the rest of the graph, or lower node $i$ degree).

This mechanism ensures the Louvain algorithm minimizes cross-community edges and cross-community nodes, achieving effective graph partitioning.

\subsection{Subgraph training}
To address subgraphs' MIS tasks, we use a \textbf{one-layer graph convolutional network (GCN)} model, detailed as follows:

For an input subgraph \( G = (V, E) \), node embeddings are initialized as random vectors. To enable differentiable approximation of discrete node selections, these initial embeddings are processed using the Gumbel-softmax trick, transforming continuous vectors into distributions that approximate discrete one-hot representations.  

The core of the model relies on a single graph convolutional layer to capture local topological dependencies. For each node \( i \), the graph convolution operation aggregates features from its neighbors \( \mathcal{N}(i) \) as:  
\begin{equation}
h_i^{(1)} = \sigma\left( \frac{1}{|\mathcal{N}(i)| + 1} \left( x_i + \sum_{j \in \mathcal{N}(i)} x_j \right) W \right)
\end{equation}  
where \( x_i \) denotes the Gumbel-softmax-processed initial embedding of node \( i \), \( W \in \mathbb{R}^{d_{\text{in}} \times d_{\text{out}}} \) is the learnable weight matrix of the convolutional layer, and \( \sigma(\cdot) \) represents a non-linear activation function (e.g., ReLU) to enhance feature representation capacity.  

After the graph convolution, a sigmoid activation is applied to the output of each node to produce the final selection probability \( p_i \) (indicating the likelihood of node \( i \) being included in the target set, such as an independent set):  
\begin{equation}
p_i = \text{sigmoid}(h_i^{(1)}) = \frac{1}{1 + \exp(-h_i^{(1)})}
\end{equation}  

Training of the model is guided by the loss function:  
\begin{align}
\mathcal{L} &= \alpha \sum_{(i,j) \in \mathcal{E}} p_i p_j - \beta \sum_{i \in \mathcal{V}} p_i
\end{align}  
Here, the first term penalizes co-selection of adjacent nodes (enforcing constraints), while the second term encourages maximizing the set size. The parameters \( \alpha \) and \( \beta \) balance these two objectives. Backpropagation is performed to update the weight matrix \( W \) iteratively until the loss \( \mathcal{L} \) stabilizes within a predefined error tolerance \( \text{lr} \), indicating convergence.

\subsection{Distributed-fine-tuning training}
Figure \ref{fig:distributed GNN training_steps} illustrates our proposed distributed-fine-tuning framework, which integrates graph partitioning, subgraph training, and reinforcement learning-based coordination. The detailed workflow is as follows:

\begin{figure*}[t]
  \centering
  \begin{subfigure}[b]{0.4\textwidth}
    \centering
    \includegraphics[width=\linewidth]{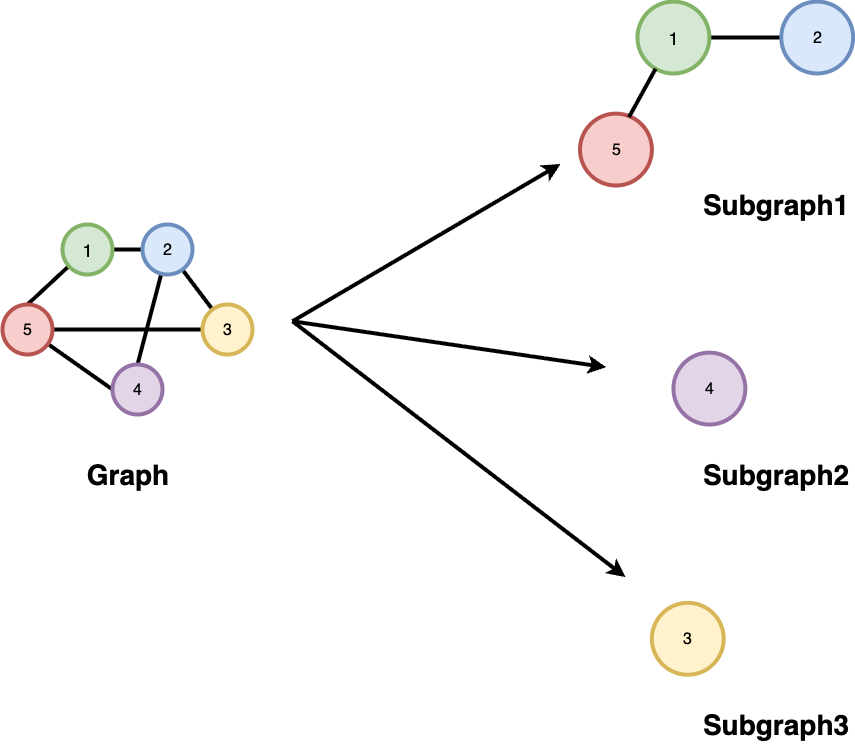}
    \caption{Graph Partitioning Step: Dividing the original graph into multiple subgraphs}
    \label{subfig:graph_partition}
  \end{subfigure}
  \hfill
  \begin{subfigure}[b]{0.5\textwidth}
    \centering
    \includegraphics[width=\linewidth]{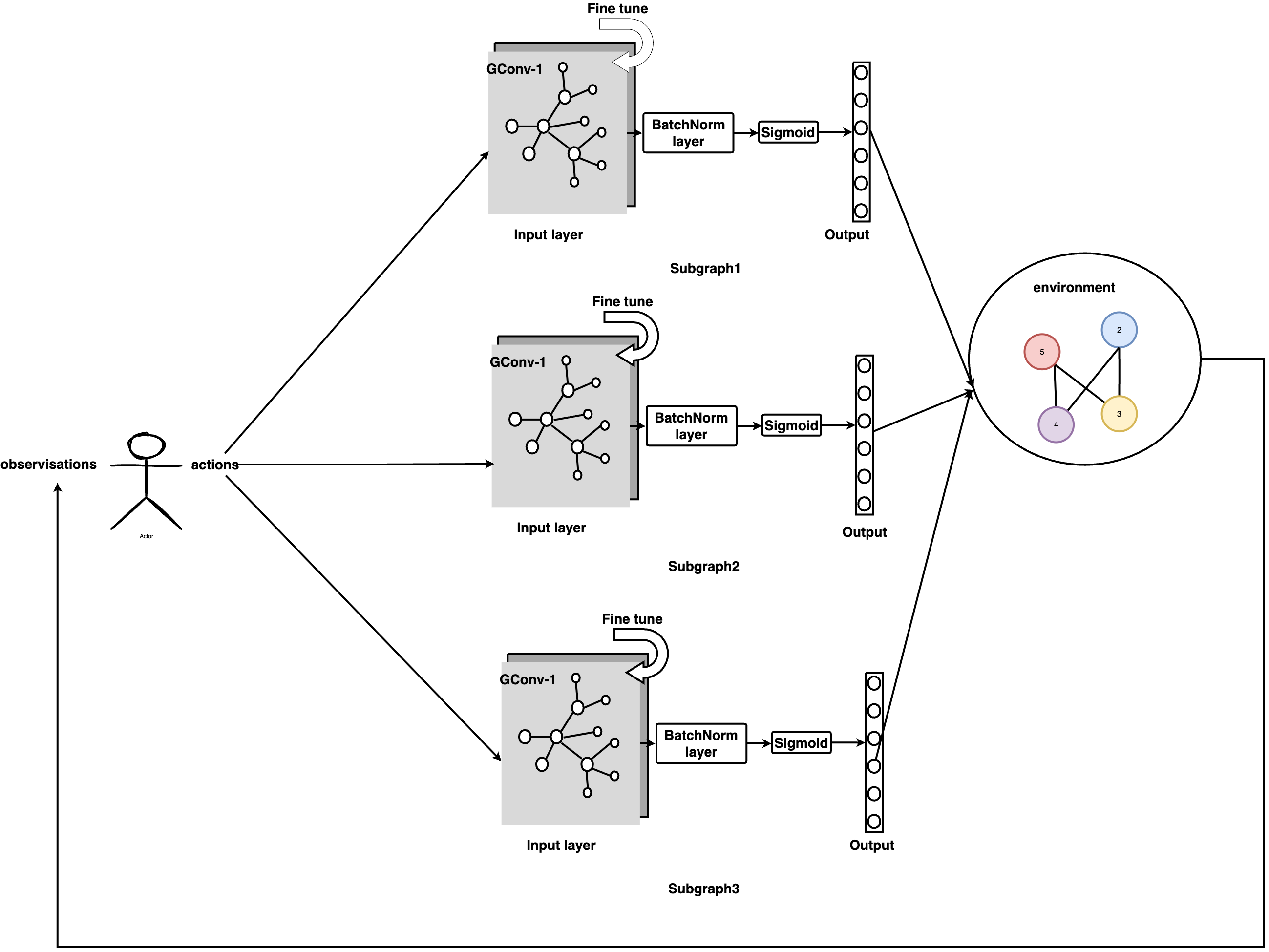}
    \caption{Reinforcement Learning (RL) Coordinated Subgraph Fine-Tuning: Optimizing subgraph models via Actor-environment interaction}
    \label{subfig:rl_finetune}
  \end{subfigure}
  \caption{Training Pipeline: (a) Graph Partitioning + (b) RL-Coordinated Subgraph Fine-Tuning}
  \label{fig:distributed GNN training_steps}
\end{figure*}


Firstly, The original large graph is partitioned into multiple subgraphs using the Louvain algorithm \cite{Blondel_Guillaume_Lambiotte_Lefebvre_2008}. Then, each subgraph is independently trained using the improved PI-GNN model until convergence, enabling local feature learning without cross-subgraph interference.
After initial training, we focus on cross-nodes (pairs of nodes $i$ and $j$ connected by an edge where $i$ and $j$ belong to different subgraphs). To minimize the QUBO value of total graph, Q-learning is employed to determine optimal fine-tuning actions. Their prediction probabilities are aggregated to form the state space for reinforcement learning:

- Cross-node probabilities $prob_{ij}$ are binarized using a 0.5 threshold ($1$ if $prob_{ij} \geq 0.5$, $0$ otherwise)

For the node $i$ and $j$, if there exists cross-edge $(i,j)$ and both $prob_i > 0.5$ and $prob_j > 0.5$, it will be identified as conflicts. To resolve conflicts, and minimize the QUBO value, we define two specific fine-tuning options for each conflicting node pair. These two options, applied across all conflict states in the graph, collectively form the action space of our reinforcement learning framework:

\begin{itemize}
    \item \textbf{Option 1:} Fine-tune the model of node $i$'s subgraph to adjust its prediction probability below the 0.5 threshold ($prob_i < 0.5$). This directly eliminates the conflict by removing node $i$ from the candidate set.
    \item \textbf{Option 2:} Fine-tune the model of node $j$'s subgraph to adjust its prediction probability below the 0.5 threshold ($prob_j < 0.5$). This resolves the conflict by excluding node $j$ from the candidate set.
\end{itemize}

Formally, the action space $\mathcal{A}$ is constructed as the union of these two options for every identified conflict state: $\mathcal{A} = \bigcup_{(i,j) \in \mathcal{C}} \{a_i^{(i,j)}, a_j^{(i,j)}\}$, where $\mathcal{C}$ denotes the set of all conflict edges, $a_i^{(i,j)}$ represents the action of fine-tuning node $i$ in conflict pair $(i,j)$, and $a_j^{(i,j)}$ represents fine-tuning node $j$ in the same pair.

The fine-tuning mechanism specifically targets the probability adjustment of conflicting nodes. For a list of selected nodes of a subgraph (e.g., $\{i_1, i_2, \dots, i_k\}$), we modify the subgraph's loss function by adding a penalty term proportional to the sum of their current prediction probabilities:
\[
\mathcal{L}_{\text{subgraph}}^{\text{new}} = \mathcal{L}_{\text{subgraph}}^{\text{original}} + \lambda \cdot \sum_{j=1}^k X_{i_j}
\]
where:
- $\mathcal{L}_{\text{subgraph}}^{\text{original}}$ is the original loss function of the subgraph,
- $\lambda$ is a positive penalty coefficient controlling the adjustment strength,
- $X_{i_j}$ represents the current prediction probability of the $j$-th selected node $i_j$ in the list (initially $> 0.5$ for conflict nodes),
- $k$ denotes the number of selected nodes in the subgraph.

This two-stage process (independent subgraph training followed by cross-node coordination) is specified in pseudocode in Algorithm~\ref{alg:rl-coordination}.

\begin{algorithm}[t]
\caption{Distributed GNN Training with RL Coordination}
\label{alg:rl-coordination}
\KwIn{Graph $G$, subgraph set $\{G_i\}$, QUBO $Q$, dataset id $d$}
\KwOut{Bitstring $\hat{x}$}
\textbf{Stage 1: Parallel Subgraph Training}\;
\ForEach{subgraph $G_i$}{
    assign device id $\gamma(i)$\;
    submit task $\mathcal{T}_i = (i, G_i, Q)$ to worker pool\;
}
\ForEach{GPU process}{
    bucket tasks by $|V_i|$ and execute sequentially\;
    store best net/optimizer states to $d\_\texttt{i}\_\texttt{nets}.pkl$\;
}
\textbf{Stage 2: Probability Reconstruction}\;
initialize $m \leftarrow \mathbf{0}^{|V|}$, $c \leftarrow \{\}$\;
\ForEach{subgraph $G_i$}{
    reload saved model; run forward to get $p_i$\;
    map local probs to global mask $m$\; 
    update cross-node mask $c$\;
}
\textbf{Stage 3: RL Coordination}\;
compute conflict edges $\mathcal{E}_\times$ from $m$\;
initialize agent base action via linear cover\;
\For{$t = 1$ to $T$}{
    select diff (explore/exploit) and derive action set $A_t$\;
    group actions per subgraph, fine-tune in parallel workers\;
    merge updated probabilities into $m_t, c_t$\;
    compute reward $r_t$; update agent Q-table\;
    keep best $(m^\star, c^\star)$ by score\;
}
return $\hat{x} \leftarrow m^\star$\;
\end{algorithm}

\section{Experiments}
\subsection{Experimental Background and Setup}
\subsubsection{Experimental Environment}
The experiment was divided into two parts: for graphs with fewer than 1,000 nodes, the artificially generated high-complexity graphs were processed on hardware with an Apple M2 processor (10 cores, 8 performance cores + 2 efficiency cores) and 16GB of unified memory. For real-world graphs, the experiments were conducted using hardware with 4 NVIDIA RTX A6000 GPUs and 49.79GB of memory. It should be noted that due to the computational performance and memory constraints of the M2 processor, the training time for some large-scale graphs in the small-node experiments may be longer than that in high-performance server environments (e.g., NVIDIA A100 GPU). However, all comparative methods were run under the same hardware configuration within each experimental part to ensure fairness in performance comparison.

\subsubsection{Selection of Comparative Methods}
We compare our methods against six baselines, including traditional heuristics and neural models: the Greedy MIS algorithm \cite{36fb03a31e54405eb79cb6307c4504b6} as a classic traditional baseline; RUN-CSP \cite{Toenshoff_Ritzert_Wolf_Grohe_2019}, a neural model designed for MIS; the PI-GNN \cite{Schuetz_Brubaker_Katzgraber_2022} with a 2-layer GCN (full-graph training); the HyperOP \cite{Heydaribeni_Zhan_Zhang_Eliassi-Rad_Koushanfar_2023} designed for hypergraph distributed training; the dynamic GNN \cite{NEURIPS2023_7fe3170d}; the GCBP\cite{Tao_Aihara_Chen_2024}; our distributed-GNN.

All methods are evaluated from three dimensions: MIS size, error rate(computed by dividing a solution’s conflictions by the MIS size), and runtime.

\subsubsection{Dataset Design}

Two types of datasets (artificially generated high-complexity graphs and real-world graphs) were used in the experiment to verify the model's performance in both "controllable complex scenarios" and "real application scenarios", as detailed below:

\begin{enumerate}

\item \textbf{Artificially Generated High-Complexity Graph Dataset:}
Regular graphs (each node has the same degree, ensuring uniform complexity of the graph structure) were generated by the function provided by the NetworkX library,
\cite{Blondel_Guillaume_Lambiotte_Lefebvre_2008}. Two variable dimensions were designed to control graph complexity, For each (node scale, degree) combination, 10 independent graph samples were generated, resulting in a total of $5 \times 4 \times 10 = 200$ graph samples. All samples were preprocessed into undirected and unweighted graphs to avoid interference from edge weights on MIS solving.
\begin{enumerate}
    \item \textbf{Node Scale}: Set to 100, 200, 500, 800, 1000 (covering small-to-medium-scale graphs to verify the model's performance change with increasing node count);
    
    \item \textbf{Average Node Degree}: Set to 10, 20, 30, 40 (higher degrees indicate denser node connections and greater difficulty in MIS solving).
\end{enumerate}

\item \textbf{Real-World Graph Dataset:}
Five classic social/academic network datasets from the Stanford Large Network Dataset Collection (SNAP) were selected \cite{snapnets}, which are facebook,CA-AstroPh, com-youtube \cite{Blondel_Guillaume_Lambiotte_Lefebvre_2008}, covering different scales and structural characteristics. All real datasets retained their original undirected graph structures.
\end{enumerate}

\subsubsection{Experimental Tasks}

The experiment focused on "Maximum Independent Set (MIS) solving" and set up 2 specific tasks to verify the effectiveness of different modules:

\begin{enumerate}

   \item  Verify the advantages of the distributed-GNN over other method, with a focus on evaluating solution quality in scenarios with "high node degrees and large-scale nodes".
   
   \item   Verify the advantages of the distributed-GNN over other method, with a focus on large-scale graph scenarios.
   
\end{enumerate}

Due to time constraints, experimental results are not recorded for graphs with fewer than 20,000 nodes where the runtime exceeds one day. 

\subsubsection{Result}

\begin{figure}[!t]
    \centering
    \includegraphics[width=\linewidth]{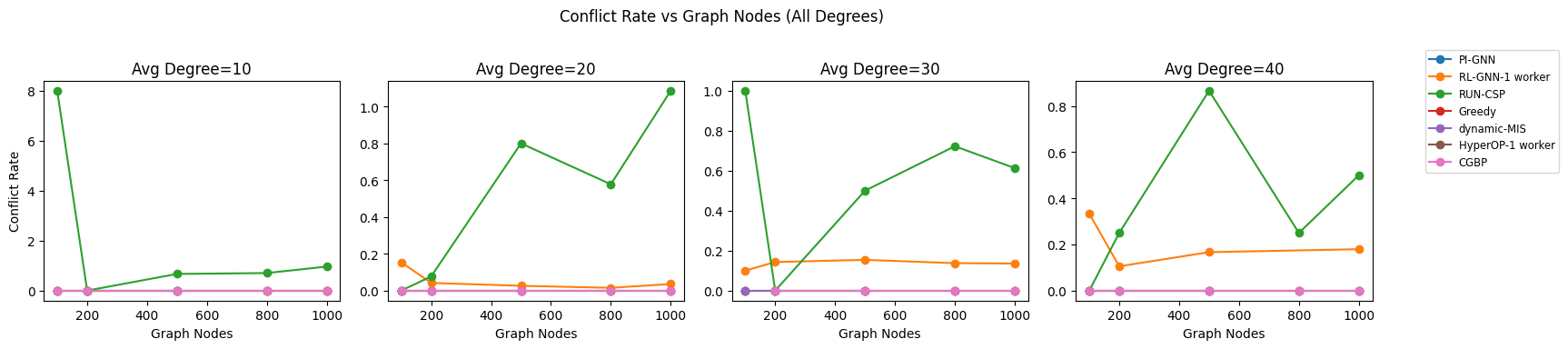}
    \caption{Conflict Rate vs Graph Nodes}
    \label{fig:conflict_rate}
\end{figure}

\begin{figure}[!t]
    \centering
    \includegraphics[width=\linewidth]{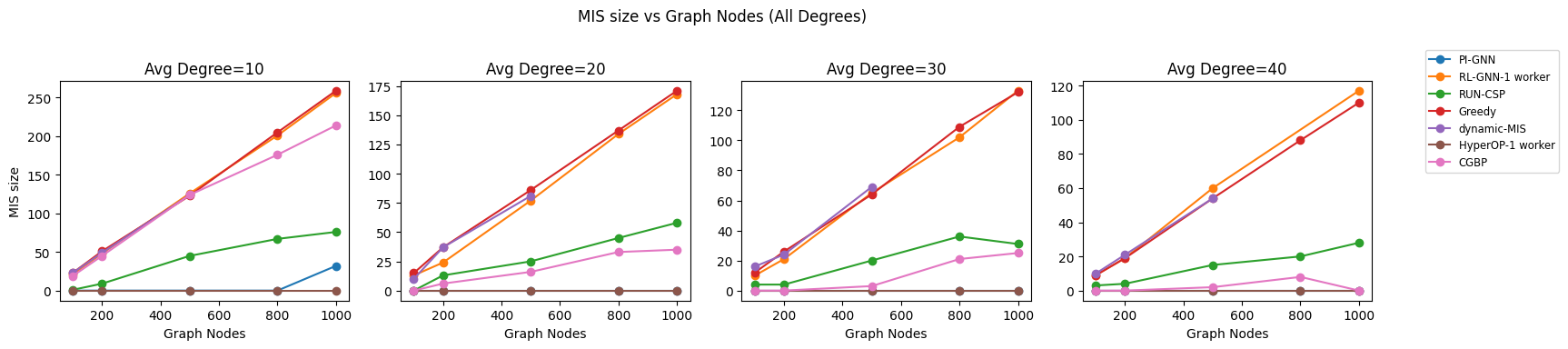}
    \caption{MIS Size vs Graph Nodes}
    \label{fig:mis_size}
\end{figure}

\begin{figure}[!t]
    \centering
    \includegraphics[width=\linewidth]{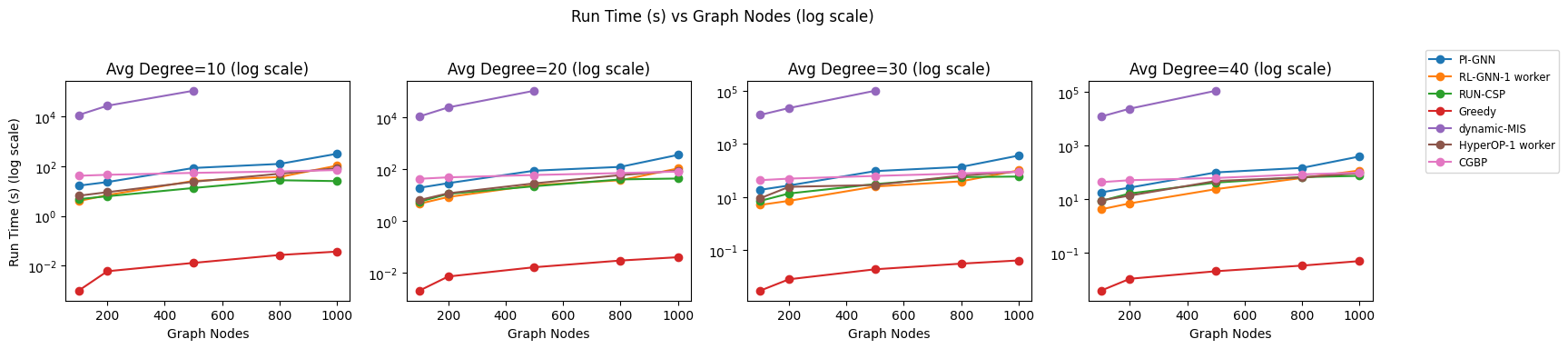}
    \caption{Run Time (s) vs Graph Nodes}
    \label{fig:runtime}
\end{figure}

\begin{table}[htbp]
    \centering
    \caption{Experimental Results of Different Methods on Various Datasets}
    \resizebox{\columnwidth}{!}{%
    \begin{tabular}{lccccccr}
        \toprule
        Method & Dataset & Edges & Nodes & Greedy & MIS & Conflict & Time (s) \\
        \midrule
        \multirow{2}{*}{PI-GNN} & Facebook & 88234 & 4039 & 1018 & 848 & 0 & 3506 \\
                                & CA-AstroPh & 198050 & 18772 & 6750 & -- & -- & OverTime \\
        \midrule
        \multirow{2}{*}{\makecell{2-workers \\ RL-GNN}} 
                                & Facebook & 88234 & 4039 & 1018 & 900 & 0.014 & 449 \\
                                & CA-AstroPh & 198050 & 18772 & 6750 & 6444 & 0.001 & 22627 \\
        \midrule
        \multirow{2}{*}{\makecell{4-workers \\ RL-GNN}} 
                                & Facebook & 88234 & 4039 & 1018 & 945 & 0.109 & 365 \\
                                & CA-AstroPh & 198050 & 18772 & 6750 & 6350 & 0.009 & 20734 \\
        \midrule
        \multirow{2}{*}{\makecell{2-workers \\ HyperOP}} 
                                & Facebook & 88234 & 4039 & 1018 & 946 & 0.012 & 36936 \\
                                & CA-AstroPh & 198050 & 18772 & 6750 & -- & -- & OverTime \\
        \midrule
        \multirow{2}{*}{\makecell{4-workers \\ HyperOP}} 
                                & Facebook & 88234 & 4039 & 1018 & 927 & 0.010 & 40173 \\
                                & CA-AstroPh & 198050 & 18772 & 6750 & -- & -- & OverTime \\
        \midrule
        \multirow{2}{*}{CGBP} & Facebook & 88234 & 4039 & 1018 & 2045 & 1.441 & 2612 \\
                                & CA-AstroPh & 198050 & 18772 & 6750 & 9908 & 0.796 & 3454 \\
        \bottomrule
    \end{tabular}}
    \label{tab:experimental_results}
\end{table}

\begin{table}[htbp]
    \centering
        \caption{Results on com-youtube Dataset}
        \resizebox{\columnwidth}{!}{%
        \begin{tabular}{lccccccr}
            \toprule
            Method & Dataset & Edges & Nodes & Greedy & MIS & Conflict & Time (s) \\
            \midrule
            \makecell{2-workers \\ RL-GNN \\ (26 subgraphs)} & com-youtube & 2987624 & 1134890 & 855607 & 869271 & 0.05 & 115890 \\
            \makecell{2-workers \\ HyperOP} & com-youtube & 2987624 & 1134890 & 855607 & -- & -- & OOM \\
            \makecell{PI-GNN} & com-youtube & 2987624 & 1134890 & 855607 & -- & -- & OOM \\
            \makecell{CGBP} & com-youtube & 2987624 & 1134890 & 855607 & -- & -- & OverTime \\
            \bottomrule
        \end{tabular}}
        \label{tab:distributed_gnn_youtube}
\end{table}

To analyze the experimental outcomes, we synthesize observations from Figures \ref{fig:conflict_rate}, \ref{fig:mis_size}, \ref{fig:runtime} and Tables \ref{tab:experimental_results}, \ref{tab:distributed_gnn_youtube} as follows:

\begin{enumerate}
    \item \textbf{Performance on High-Complexity Graphs:}
    As illustrated in Figures \ref{fig:conflict_rate}–\ref{fig:runtime}, the comparative results reveal distinct performance patterns among different methods under varying average degrees and node scales.

    \begin{itemize}
        \item \textbf{HyperOP and PI-GNN:} As the average node degree increases (i.e., as graph complexity grows), both methods exhibit a pronounced decline in solution quality. In particular, when the average degree exceeds 20 and the graph scale becomes large, their performance in conflict-rate control and Maximum Independent Set (MIS) size optimization drops sharply, and in some cases, they fail to produce valid solutions. This indicates their limited adaptability to highly complex graph structures.
        
        \item \textbf{RUN-CSP:} Although RUN-CSP can solve partial subsets of the MIS, its solution quality deteriorates significantly with increasing graph complexity. As both average degree and node count grow, the number of edge conflicts rises substantially, leading to reduced overall solution effectiveness in high-complexity scenarios.
        
        \item \textbf{Dynamic-MIS:} While achieving competitive MIS sizes, Dynamic-MIS suffers from scalability issues—its runtime increases almost linearly with graph complexity, and memory consumption grows dramatically, often exceeding hardware limits even on graphs with roughly 1,000 nodes. Consequently, it becomes impractical for large-scale, high-density graphs due to prohibitive computational and memory costs.

        \item \textbf{RL-GNN:} Across all high-complexity scenarios, the proposed RL-GNN achieves high-quality solutions within relatively short runtimes. It maintains competitive MIS sizes while effectively controlling conflict rates, demonstrating both efficiency and robustness in handling complex graph structures.

        \item \textbf{CGBP:} Although CGBP shows advantages in runtime efficiency and CGBP exhibits substantial improvement over PI-GNN when dealing with graphs of moderate density (average node degree $\approx$10). However, as graph connectivity continues to increase, the MIS size achieved by CGBP declines sharply, suggesting that the model struggles to maintain solution quality under high-complexity graph structures.
    \end{itemize}

    \item \textbf{Performance on Large-Scale Graphs:}
    The results summarized in Tables \ref{tab:experimental_results} and \ref{tab:distributed_gnn_youtube} further highlight scalability trends across different methods.
    
    \begin{itemize}
        \item \textbf{RL-GNN:} As the number of workers increases (e.g., from 2 to 4), a slight rise in conflict rate can be observed (for instance, on the Facebook dataset, from 0.014 with 2 workers to 0.109 with 4). Nonetheless, results on the com-Youtube dataset (1,134,890 nodes) confirm that the proposed method successfully solves the MIS problem on ultra-large-scale graphs, demonstrating excellent scalability and effectiveness.
        \item \textbf{HyperOP:} During distributed training, HyperOP requires iterative aggregation of subgraph solutions across all devices, leading to substantial communication overhead. As a result, its total runtime increases significantly—for example, on the Facebook dataset, the 2-worker HyperOP configuration requires 36,936 seconds, compared to only 4,004 seconds for PI-GNN under the same conditions—highlighting its inefficiency in distributed communication and scalability.
        \item \textbf{CGBP:} Although CGBP demonstrates significant advantages in terms of computational efficiency compared with other methods, its solution quality remains inferior, as the obtained solution set contains numerous conflicts
        \item \textbf{PI-GNN:} The traditional PI-GNN method fails to run on large-scale datasets due to memory limitations.
    \end{itemize}
\end{enumerate}

\section{Conclusion}
This study focuses on improving the performance of applying Graph Neural Networks (GNNs) to the combinatorial optimization problem, including scalability and solution quality for complex graphs. Acoording to section IV experiments, the performance of model on time cost and solution quality can be summarized as follows.

First, the model shows improved ability in handling high-complexity graphs, which still can obtain validate solutions for the graph with high average degree, comparing to other neural network methods, its solution has lower conflict rate and larger independent set size.  

Second, the distributed training design enables the model can process the graph which can not be trained in a single device because of the memory limits, on the result of that, it explore the neural network such as GNN-based capability of applying to larger problem instances. However, there are still several limitations. 
if there are plenty of conflicts arise after subgrpahs full-training finished, the reinforcement learning (RL) coordination will spend a lot of time on searching optimal solution, which can resolve conflicts and find the maximum independent size of whole graph, 
and solution quality may decline compared to single node training, because, in fine-tune processing, the subgraph training model may lose part of validate solution set. Moreover, subgraph partitions quality has effective influences on the output of the distributed model. when cross-subgraph nodes accounts for a high proportion of all nodes, the time spent on RL finding optimal solution may exceed that on full-graph training. 
For the graph can be trained in a single node, the advantages of distributed training in time cost will be disappered, because of the overhead of RL coordination mechanism.
Additionally, the convergence of the RL-based coordination mechanism lacks formal theoretical proof, which adds another potential risks to optimization stability.

To address these challenges, several existing research directions can be taken into account. One is to optimize the process of subgraph fine-tune, there are several adaptive optimization strategies proposed, for example, 
The fine-tuning mechanism can be designed to focus on the region with high conflictions, taking mini batch to fine-tune the subgraph, to avoid impacting on other unrelated nodes, which can also reduce unnecessary iterations and achieve a less time cost.

Another promising method can be considered is adapting GNN pre-training strategies. GNN can quickly shift between excluding node and including node scenarios through pre-training, thereby optimizing the overfitting and underfitting situations resulting from the subgraph fine-tuning and improving adaptability to dynamic graphs where nodes and edges change over time \cite{Hu_Liu_Gomes_Zitnik_Liang_Pande_Leskovec_2019, Sun_2023}. 

In the future, a more ambitious direction lies in developing a pipeline structure including natural data input, encode model, which can transfer natural data to learnable features in GNN, and then the combinatorial optimization solving network can processes these features, finally, the structure will decode the output features of nodes to obtain solution for real-world problems. 
Such end-to-end architectures would extend existing traditional method, it can not only to learn constraints from manual processed data but also to formalize and reason within real-world data, and build graph structures through natural data.  

Overall, this work advances the state of GNN-based MIS solving for complex graphs and validates distributed computation as a practical and scalable approach, providing another path to tackle large-scale combinatorial optimization challenges.

\printbibliography

@article{Schuetz_Brubaker_Katzgraber_2022,
  title = {Combinatorial optimization with physics-inspired graph neural networks},
  volume = {4},
  url = {https://doi.org/10.1038/s42256-022-00468-6},
  DOI = {10.1038/s42256-022-00468-6},
  number = {4},
  journal = {Nature Machine Intelligence},
  publisher = {Springer Science and Business Media LLC},
  author = {Schuetz, Martin J. A. and Brubaker, J. Kyle and Katzgraber, Helmut G.},
  year = {2022},
  month = apr,
  pages = {367--377}
}

@article{Johnson2022,
  author = {Johnson, B. and others},
  title = {Efficient Vehicle Routing in Dynamic Logistics Networks},
  journal = {Transportation Research Part E: Logistics and Transportation Review},
  year = {2022},
  volume = {165},
  pages = {102856}
}

@article{Lee2021,
  author = {Lee, C. and others},
  title = {Next-Generation Chip Design with Combinatorial Optimization},
  journal = {IEEE Transactions on Semiconductor Manufacturing},
  year = {2021},
  volume = {34},
  number = {4},
  pages = {485--493}
}

@article{Wang2020,
  author = {Wang, D. and others},
  title = {Enhancing User Experience through Recommendation System Optimization},
  journal = {ACM Transactions on Internet Technology},
  year = {2020},
  volume = {20},
  number = {3},
  pages = {1--20}
}

@article{Brown2023,
  author = {Brown, E. and others},
  title = {Smart Grid Load Management Using Combinatorial Optimization},
  journal = {IEEE Transactions on Smart Grid},
  year = {2023},
  volume = {14},
  number = {3},
  pages = {1856--1867}
}

@article{Liu2024,
  author = {Liu, X. and others},
  title = {Novel Strategies for Combinatorial Optimization in the Era of Big Data and Complex Systems},
  journal = {Journal of Advanced Computational Intelligence},
  year = {2024},
  volume = {30},
  number = {2},
  pages = {187--201}
}

@article{Cappart2021,
  title = {Combinatorial optimization and reasoning with graph neural networks},
  journal = {arXiv:2102.09544},
  url = {https://arxiv.org/abs/2102.09544},
  author = {Cappart, Quentin and Ch{\'e}telat, Didier and Khalil, Elias and Lodi, Andrea and Morris, Christopher and Veli{\v{c}}kovi{\'c}, Petar},
  year = {2021}
}

@article{Blondel_Guillaume_Lambiotte_Lefebvre_2008,
  title = {Fast unfolding of communities in large networks},
  volume = {2008},
  url = {https://doi.org/10.1088/1742-5468/2008/10/p10008},
  DOI = {10.1088/1742-5468/2008/10/p10008},
  number = {10},
  journal = {Journal of Statistical Mechanics: Theory and Experiment},
  publisher = {IOP Publishing},
  author = {Blondel, Vincent D and Guillaume, Jean-Loup and Lambiotte, Renaud and Lefebvre, Etienne},
  year = {2008},
  month = oct,
  pages = {P10008}
}

@article{Toenshoff_Ritzert_Wolf_Grohe_2019,
  title = {Graph Neural Networks for Maximum Constraint Satisfaction},
  journal = {arXiv:1909.08387},
  url = {https://arxiv.org/abs/1909.08387},
  author = {Toenshoff, Jan and Ritzert, Martin and Wolf, Hinrikus and Grohe, Martin},
  year = {2019}
}

@misc{snapnets,
  author = {Jure Leskovec and Andrej Krevl},
  title = {{SNAP Datasets}: {Stanford} Large Network Dataset Collection},
  howpublished = {\url{http://snap.stanford.edu/data}},
  month = jun,
  year = 2014
}

@article{Heydaribeni_Zhan_Zhang_Eliassi-Rad_Koushanfar_2023,
  title = {Distributed Constrained Combinatorial Optimization leveraging Hypergraph Neural Networks},
  journal = {arXiv:2311.09375},
  url = {https://arxiv.org/abs/2311.09375},
  author = {Heydaribeni, Nasimeh and Zhan, Xinrui and Zhang, Ruisi and Eliassi-Rad, Tina and Koushanfar, Farinaz},
  year = {2023}
}

@inproceedings{NEURIPS2023_7fe3170d,
  author = {Brusca, Lorenzo and Quaedvlieg, Lars C.P.M. and Skoulakis, Stratis and Chrysos, Grigorios and Cevher, Volkan},
  booktitle = {Advances in Neural Information Processing Systems},
  editor = {A. Oh and T. Naumann and A. Globerson and K. Saenko and M. Hardt and S. Levine},
  pages = {40637--40658},
  publisher = {Curran Associates, Inc.},
  title = {Maximum Independent Set: Self-Training through Dynamic Programming},
  url = {https://proceedings.neurips.cc/paper_files/paper/2023/file/7fe3170d88a8310ca86df2843f54236c-Paper-Conference.pdf},
  volume = {36},
  year = {2023}
}

@article{Gamarnik_2023,
  title = {Barriers for the performance of graph neural networks (GNN) in discrete random structures},
  journal = {arXiv:2306.02555},
  url = {https://arxiv.org/abs/2306.02555},
  author = {Gamarnik, David},
  year = {2023}
}

@article{Peng_Choi_Xu_2020,
  title = {Graph Learning for Combinatorial Optimization: A Survey of State-of-the-Art},
  journal = {arXiv:2008.12646},
  url = {https://arxiv.org/abs/2008.12646},
  author = {Peng, Yun and Choi, Byron and Xu, Jianliang},
  year = {2020}
}

@article{Hu_Liu_Gomes_Zitnik_Liang_Pande_Leskovec_2019,
  title = {Strategies for Pre-training Graph Neural Networks},
  journal = {arXiv:1905.12265},
  url = {https://arxiv.org/abs/1905.12265},
  author = {Hu, Weihua and Liu, Bowen and Gomes, Joseph and Zitnik, Marinka and Liang, Percy and Pande, Vijay and Leskovec, Jure},
  year = {2019}
}

@article{Sun_2023,
  title = {Pp-Gnn: Pretraining Position-Aware Graph Neural Networks with the Np-Hard Metric Dimension Problem},
  url = {https://doi.org/10.2139/ssrn.4453304},
  DOI = {10.2139/ssrn.4453304},
  publisher = {Elsevier BV},
  author = {Sun, Michael},
  year = {2023}
}

@article{Darvariu_Hailes_Musolesi_2024,
  title = {Graph Reinforcement Learning for Combinatorial Optimization: A Survey and Unifying Perspective},
  journal = {arXiv:2404.06492},
  url = {https://arxiv.org/abs/2404.06492},
  author = {Darvariu, Victor-Alexandru and Hailes, Stephen and Musolesi, Mirco},
  year = {2024}
}

@article{Meirom_Maron_Mannor_Chechik_2020,
  title = {Controlling Graph Dynamics with Reinforcement Learning and Graph Neural Networks},
  journal = {arXiv:2010.05313},
  url = {https://arxiv.org/abs/2010.05313},
  author = {Meirom, Eli A. and Maron, Haggai and Mannor, Shie and Chechik, Gal},
  year = {2020}
}

@inbook{36fb03a31e54405eb79cb6307c4504b6,
title = "The differential equation method for random graph processes and greedy algorithms",
author = "Wormald, {Nicholas Charles}",
year = "1999",
language = "English",
pages = "73--155",
booktitle = "Lectures on Approximation and Randomized Algorithms",
publisher = "Wydawnictwo Naukowe Pwn",
}

@article{Vinyals_Fortunato_Jaitly_2015, title={Pointer Networks}, volume={arXiv:1506.03134}, archiveLocation={1506.03134}, url={https://arxiv.org/abs/1506.03134}, journal={arXiv preprint}, publisher={arXiv}, author={Vinyals, Oriol and Fortunato, Meire and Jaitly, Navdeep}, year={2015} }

@article{Tao_Aihara_Chen_2024, title={Brain-inspired Chaotic Graph Backpropagation for Large-scale Combinatorial Optimization}, volume={arXiv:2412.09860}, archiveLocation={2412.09860}, url={https://arxiv.org/abs/2412.09860}, journal={arXiv preprint}, publisher={arXiv}, author={Tao, Peng and Aihara, Kazuyuki and Chen, Luonan}, year={2024} }

@article{Bello_Pham_Le_Norouzi_Bengio_2016, title={Neural Combinatorial Optimization with Reinforcement Learning}, volume={arXiv:1611.09940}, archiveLocation={1611.09940}, url={https://arxiv.org/abs/1611.09940}, journal={arXiv preprint}, publisher={arXiv}, author={Bello, Irwan and Pham, Hieu and Le, Quoc V. and Norouzi, Mohammad and Bengio, Samy}, year={2016} }

@article{Kool_Hoof_Welling_2018, title={Attention, Learn to Solve Routing Problems!}, volume={arXiv:1803.08475}, archiveLocation={1803.08475}, url={https://arxiv.org/abs/1803.08475}, journal={arXiv preprint}, publisher={arXiv}, author={Kool, Wouter and Hoof, Herke van and Welling, Max}, year={2018} }

@inproceedings {273707,
author = {Swapnil Gandhi and Anand Padmanabha Iyer},
title = {P3: Distributed Deep Graph Learning at Scale},
booktitle = {15th {USENIX} Symposium on Operating Systems Design and Implementation ({OSDI} 21)},
year = {2021},
isbn = {978-1-939133-22-9},
pages = {551--568},
url = {https://www.usenix.org/conference/osdi21/presentation/gandhi},
publisher = {{USENIX} Association},
month = jul
}

@article{Harlap_Narayanan_Phanishayee_Seshadri_Devanur_Ganger_Gibbons_2018, title={PipeDream: Fast and Efficient Pipeline Parallel DNN Training}, volume={arXiv:1806.03377}, archiveLocation={1806.03377}, url={https://arxiv.org/abs/1806.03377}, journal={arXiv preprint}, publisher={arXiv}, author={Harlap, Aaron and Narayanan, Deepak and Phanishayee, Amar and Seshadri, Vivek and Devanur, Nikhil and Ganger, Greg and Gibbons, Phil}, year={2018} }

@article{Niu_Recht_Re_Wright_2011, title={HOGWILD!: A Lock-Free Approach to Parallelizing Stochastic Gradient Descent}, volume={arXiv:1106.5730}, archiveLocation={1106.5730}, url={https://arxiv.org/abs/1106.5730}, journal={arXiv preprint}, publisher={arXiv}, author={Niu, Feng and Recht, Benjamin and Re, Christopher and Wright, Stephen J.}, year={2011} }

@article{Tahami2022,
  author       = {H.\ Tahami},
  title        = {A Literature Review on Combining Heuristics and Exact Algorithms in Combinatorial Optimization},
  journal      = {International Journal of Computational Engineering Science},
  year         = {2022},
  volume       = {3},
  number       = {1},
  pages        = {1--32},
  note         = {Survey paper including discussion of dynamic programming and integer programming as exact methods},
}

@inproceedings{Raidl2008,
  author       = {G.~R. Raidl and J. Puchinger},
  title        = {Combining (Integer) Linear Programming Techniques and Metaheuristics},
  booktitle    = {Proceedings First International Work-Conference on the Interplay Between Natural and Artificial Computation},
  year         = {2008},
  pages        = {3--22},
  note         = {Discusses integer programming, metaheuristics including genetic algorithms and simulated annealing},
}

\end{document}